# BPLight-CNN: A Photonics-based Backpropagation Accelerator for Deep Learning

DHARANIDHAR DANG, UC San Diego
SAI VINEEL REDDY CHITTAMURU, Micron Technology
SUDEEP PASRICHA, Colorado State University
RABI MAHAPATRA, Texas A&M University
DEBASHIS SAHOO, UC San Diego

Training deep learning networks involves continuous weight updates across the various layers of the deep network while using a backpropagation algorithm (BP). This results in expensive computation overheads during training. Consequently, most deep learning accelerators today employ pre-trained weights and focus only on improving the design of the inference phase. The recent trend is to build a complete deep learning accelerator by incorporating the training module. Such efforts require an ultra-fast chip architecture for executing the BP algorithm. In this article, we propose a novel photonics-based backpropagation accelerator for high performance deep learning training. We present the design for a convolutional neural network, *BPLight-CNN*, which incorporates the silicon photonics-based backpropagation accelerator. *BPLight-CNN* is a first-of-its-kind photonic and memristor-based CNN architecture for end-to-end training and prediction. We evaluate *BPLight-CNN* using a photonic CAD framework (IPKISS) on deep learning benchmark models including LeNet and VGG-Net. The proposed design achieves (i) at least 34× speedup, 34× improvement in computational efficiency, and 38.5× energy savings, during training; and (ii) 29× speedup, 31× improvement in computational efficiency, and 38.7× improvement in energy savings, during inference compared to the state-of-the-art designs. All these comparisons are done at a 16-bit resolution; and BPLight-CNN achieves these improvements at a cost of approximately 6% lower accuracy compared to the state-of-the-art.

## 1. INTRODUCTION

In today's era of big data, the volume of data that computing systems process has been increasing exponentially. Deep neural networks have become the state-of-the-art across a broad range of big data applications such as speech processing, image recognition, financial predictions, etc. Convolutional neural networks (CNNs) are a popular deep learning framework with superior accuracy on applications that deal with videos and images. However, CNNs are highly compute and memory intensive, requiring enormous computational resources. With Moore's law coming to an end, traditional Von Neuman computing systems such as heterogeneous CPU/GPU platforms cannot address this high computational demand, within reasonable power and processing time limitations. Therefore, several FPGA [1] and ASIC [2] approaches have been proposed to accomplish large-scale deep learning acceleration.

A CNN comprises of two stages: training and inference (i.e., testing). Most hardware accelerators for CNNs in prior literature focus only on the inference stage. However, training a CNN is several hundred times more compute intensive and power intensive than its inference [3].

Authors' addresses: D. Dang (corresponding author) and D Sahoo, Department of Computer Science & Engineering, UC San Diego, CA 92093. S. V. R. Chittamuru, Micron Technology, Inc., Austin, Texas 78728. S. Pasricha, Department of Electrical and Computer Engineering, Colorado State University, Fort Collins, CO, 80523; R. Mahapatra, Department of Computer Science and Engineering, Texas A&M University, College Station, Texas 77843; email: {dhdang, dsahoo}@health.ucsd.edu, chittamuru@micron.com, sudeep @colostate.edu, rabi@tamu.edu.

Moreover, for many applications, training is not just a one-time activity, especially under changing environmental and system conditions, where re-training of the CNN at regular intervals is essential to maintaining prediction accuracy for the application over time.

Training a CNN in general, incorporates a backpropagation algorithm which involves notable memory locality and compute parallelism. Recently, a few resistive memory (memristor) based training accelerators have been demonstrated for CNNs, e.g. ISAAC [2], PipeLayer [4], RCP [5], and MNN [6]. ISAAC, RCP and MNN use highly parallel memristor crossbar arrays to address the need for parallel computations in CNNs. In addition, ISAAC uses a very deep pipeline to improve system throughput. However, this is only beneficial when a large number of consecutive images can be fed into the architecture. Unfortunately, during training, in many cases, a limited number of consecutive images need to be processed before weight updates. The deep pipeline in ISAAC also introduces frequent pipeline bubbles. Compared to ISAAC, PipeLayer demonstrates an improved pipeline approach to enhance throughput. However, RCP, MNN, ISAAC, and PipeLayer involve several analog-to-digital (AD) and digital-to-analog (DA) conversions which become a performance bottleneck, in addition to their large power consumption. Also, training in these accelerators involves sequential weight updates from one layer to another. This incurs inter-layer waiting time for synchronization, which reduces overall performance. This motivates an analog accelerator that can drastically reduce the number of AD/DA conversions, and inter-layer waiting time.

It has been recently demonstrated that a completely analog matrix-vector multiplication is 100× more efficient than its digital counterpart implemented with an ASIC, FPGA, or GPU [7]. HP labs have showcased a memristor dot product engine that can achieve a speed-efficiency product of 1000× compared to a digital ASIC [7]. The high efficiency of analog computing with memristors motivates their usage in the construction of next generation accelerators. Moreover, several prior works have emphasized the importance of using ultrafast photonic devices in hardware accelerators for CNNs. For example, matrix multiplication based on photonic devices is demonstrated in [9]-[10]. Photonic device-based training with backpropagation was proposed in [11]. Vandroome et al. in [8] have demonstrated a small-scale efficient recurrent neural network using analog photonic computing. A few efficient on-chip photonic accelerators have also been proposed in [12]-[17], but all of these accelerators primarily focus on inference only. Furthermore, the CNN accelerators proposed in prior works [12]-[13] are promising for small-scale CNNs, but for large-scale CNNs, data in each layer needs to be stored back in a DRAM, which limits their performance and scalability for universal adoption. [14] demonstrates an inference small-scale CNN accelerator using silicon photonics. A memristor integrated photonic design for medium scale inference CNN accelerator is proposed in [15]. [16] adopts a hybrid approach in which convolution is performed using silicon photonics circuitry and other operations are performed using digital ALUs. This leads to multiple inter-layer O/E and A/D conversions followed by E/O conversions. As a result, although promising, the performance is limited. In [17], the authors have demonstrated a photonic inference only CNN accelerator. They have shown a few orders of magnitude improvement in speedup and energy-efficiency for binarized CNN accelerators. It is not clear if such a design could be applied to generic CNN accelerator design. In summary, a full-fledged analog CNN accelerator that is capable of both training and inference has yet to be demonstrated.

In this article, we propose a novel silicon photonics-based backpropagation accelerator for training CNNs. We present the design of this novel CNN accelerator (*BPLight-CNN*) that integrates the photonics-based backpropagation accelerator. *BPLight-CNN* is a first-of-its-kind memristor-integrated silicon photonic CNN accelerator for end-to-end training and inference. It is intended to perform highly energy efficient and ultra-fast training for deep learning applications with state-of-the-art prediction accuracy. The main contributions of this article are summarized as follows:

— We propose *BPLight-CNN*, a fully analog and scalable silicon photonics-based backpropagation accelerator in conjunction with a configurable memristor-integrated photonic CNN accelerator design;
— We demonstrate a pipelined data distribution approach for high throughput training with *BPLight-CNN*;

— We synthesize the *BPLight-CNN* architecture using a photonic CAD framework (i.e., IPKISS). The synthesized *BPLight-CNN* is used to execute four variants of VGG-Net and two variants of LeNet, demonstrating at least 35×, 31×, and 45× improvements in throughput, computation efficiency, and energy efficiency, respectively, compared to the state-of-the-art CNN accelerators. All these comparisons are done at a 16-bit resolution; and *BPLight-CNN* achieves these improvements at a cost of approximately 6% lower accuracy compared to the state-of-the-art. For 32-bit resolution *BPLight-CNN* attains state-of-the-art accuracy with negligible effects on energy savings and speedup.

The rest of the article is organized as follows. Section 2 introduces on-chip photonic components which are the building blocks of the BPLight-CNN architecture. Section 3 presents a brief overview of CNNs. Section 4 provides a detailed description of the *BPLight-CNN* architecture. Section 5 illustrates an implementation of a popular CNN model using the proposed *BPLight-CNN* architecture. Section 6 presents the experimental setup, results, and comparative analysis. Lastly, we present concluding remarks in Section 7.

## 2. ON-CHIP PHOTONICS COMPONENTS: OVERVIEW

The BPLight-CNN architecture is a fully analog photonic-based accelerator. To understand this architecture, in this section we introduce some of the basic on-chip photonics components that are utilized by it, such as photonic waveguides, microring modulators (MRMs), semiconductor-optical-amplifiers (SOAs), photodetectors, and multi-wavelength laser sources for on-chip photonic signaling [18].

An MRM is a circular shaped photonic structure with a radius of ~5 μm which is used to modulate electronic signals onto a photonic signal at the transmission source in a waveguide. MRMs are also used to couple/filter out light from the waveguide at the destination. Each MRM modulates/couple light of a specific wavelength. The geometry of the MRM determines its wavelength selectivity. We can also inject (remove) charge carriers to (from) an MRM to alter its operating wavelength. An SOA is an optoelectronic device that under suitable operating conditions can amplify photonic signals. A detailed description of the structure, functionality, and modeling of SOAs is given in [19].

In a typical high bandwidth photonic link, an off-chip laser source (either on the board or on a 2.5D interposer) generates multiple wavelengths, which are coupled by an optical grating coupler to an on-chip photonic waveguide. The use of multiple wavelengths (e.g., 32) to transmit multiple streams of bits simultaneously is referred to as dense-wavelength-division-multiplexing (DWDM). To enable processing of these photonic signals, the on-chip photonic waveguide guides the input optical power of these DWDM photonic signals via a series of MRMs (where each MRM operates on a photonic signal with specific wavelength) and SOAs. Finally, the photonic signals arrive at the destination where they are coupled out of the waveguide by MRMs, which drop the photonic signals onto photodetectors, to convert them back to electronic signals.

An important characteristic of photonic signal transmission in an on-chip photonic link is that it is inherently lossy, i.e., the photonic signal is subject to losses such as insertion losses in MRMs, active region losses in SOAs, detection losses in photo-detectors, and propagation and bending losses in waveguides. In addition, there are splitting and coupling losses in grating couplers, splitters, and multiplexers. Higher laser power is needed to compensate for the losses, for reliable photonic signal transmission. These photonic links are used to construct parts of our *BPLight-CNN* architecture, as discussed next.

## 3. CONVOLUTIONAL NEURAL NETWORKS: OVERVIEW

### 3.1 Basics of Convolutional Neural Network

Convolutional neural networks (CNNs) are a class of feed-forward neural networks commonly used for analyzing visual imagery for image classification and object detection/prediction tasks. CNNs comprise of a sequence of hidden layers where each layer is composed of neurons arranged in three dimensions: width, height and number of channels. The neurons in a layer are connected to a small region of the layer before it. This ensures that weights are shared among the neural connections across adjacent layers, thereby reducing the number of parameters (weights) to be learnt in the network. The final output layer in a CNN is a fully connected neural network (FCN) that transforms the full input image into a single vector of class scores arranged along the channel dimension.

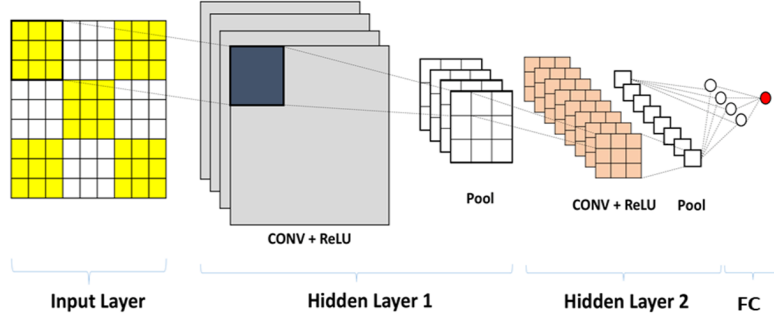

Figure 1: Overview of CNN with two hidden layers and an FC layer. Each hidden layer comprises of [CONV-POOL].

In principle, three types of layers are used to build a CNN: convolution layer (CONV), pooling layer (POOL) and a fully connected layer (FC). Generally, CONV is accompanied with a non-linear activation function, such as ReLU. Depending on the sequence in which these layers are arranged, there are different CNN models, such as AlexNet [40], VGG [42], LeNet [43], GoogLeNet [33] etc. Furthermore, LeNet has lesser number of kernels compared to VGG (see Table 1). Therefore, VGG is capable of processing large images with more features with high accuracy, whereas the simpler LeNet can provide high accuracy for inputs with small images that require less features to be learned. Fig. 1 illustrates an example of a CNN with [CONV-POOL]-[CONV-POOL]-[FC] i.e., 2 hidden layers each of which comprises of [CONV-POOL]. LeNet has the configuration [CONV-POOL]-[CONV-POOL]-[2FC] and VGG16 is built with [2CONV-POOL]-[2CONV-POOL]-[3CONV-POOL]-[3CONV-POOL]-[3CONV-POOL]-[3FC].

The functional details of the various layers are as follows.

1) **Convolution layer** (CONV) is used to extract features from the image using multiple filters. An $M \times N$ CONV receives M features as input and produces N features as output. It uses a set of M filters (or kernels), each of size $F_1 \times F_2$. Each of these filters slide across a corresponding feature with a stride of $S_1 \times S_2$ to perform element-wise vector matrix multiplication. The resulting N output features can be represented using the following equation:

$$Out[n][p][q] = \sum_{m=0}^{M} \sum_{i=0}^{F_1} W(n,p,q) \qquad (1)$$

where,

$$W(n,p,q) = \sum_{j=0}^{F_2}\big[W[n][m][i][j] \times In[S_1 * p + i][S_2 * q + j]\big] \qquad (2)$$

Here, *n* and *m* are kernel index, *i* and *j* are (x,y) values of a kernel, and *p* and *q* are (x,y) values of input *In*.

2) **Neural activation layer** performs a biological activation function such as a sigmoid, rectified linear unit (ReLU), or tanh, on each feature of its previous layer. We utilize ReLU which is a widely used non-linear activation function with state-of-the-art performance [1], [7], [40], which can be described as follows:

$$ReLU(x) = \max\{0, x\} \tag{3}$$

3) **Pooling layer** (POOL) is used to obtain spatial invariance while scaling features from preceding layers. A 'maximum/average of many features' approach is considered to scale down extracted features. POOL maintains translational invariance, or in other words, it results in a scaled-down feature map identical to its original version.

4) **Fully Connected Network layer** (FC) performs the final classification or prediction in a CNN. An FC takes the feature maps generated from previous layers and multiplies a weight matrix following a dense matrix vector multiplication pattern. A few cascaded FC layers carry out the same procedure to produce the final classification or prediction output. The computation of an FC layer can be described by the equation:

$$Out[p][q] = \sum_{n=0}^{N}[W[p][n] \times In[n][q]] \tag{4}$$

## 3.2 Backpropagation Algorithm

A deep neural network such as a CNN has two stages: training and inference (testing). In the training phase, the filter weights (and biases) in CONV and FC layers are learnt by using a backpropagation (BP) algorithm. The BP algorithm involves a forward and a backward pass in the deep network. Given a training sample $x$ in the forward pass, the weighted input sum (convolution) $z$ is computed for neurons in each layer $l$ with some initial filter weights $w$ (and bias $b$) followed by neural activation $\sigma(z)$ (ReLU($z$) in our work), and POOL. The final layer $L$ computes the output label of the overall network for every forward pass. This can be summarized as follows:

**Forward Pass:** For each layer $l$,

$$z^{x,l} \leftarrow w^l a^{x,l-1} + b^l \tag{5}$$

$$a^{x,l} \leftarrow \sigma(z^{x,l}) \tag{6}$$

A cost function $C$ is defined to quantitatively evaluate how well the output of a neural network at the final layer $L$ compares to the target class label. The optimization goal in training is to minimize this cost function. The output error in the final prediction $\delta^{x,L}$ is a result of errors induced by the neurons in each hidden layer during the forward pass. To compute the error contribution of a neuron in the previous layer i.e., $\delta^{x,l}$, the final error is back propagated through the network starting from the output layer. This can be summarized as follows:

**Output error:** At the final layer $L$,

$$\delta^{x,L} \leftarrow \nabla_a C_x \odot \sigma'(z^{x,L}) \tag{7}$$

**Backward Pass:** For each layer $l$,

$$\delta^{x,l} \leftarrow ((w^{l+1})^T \times \delta^{x,l+1}) \odot \sigma'(z^{x,l}) \tag{8}$$

Here, $\nabla_a$ is gradient of $a^{x,l}$, $\odot$ is the dot product, and $\sigma'(z^{x,L})$ is derivative of $\sigma(z^{x,L})$. These error contributions are necessary to update the filter weights $w$ and biases $b$ in the respective layers using a gradient descent method. In gradient descent, the forward and backward pass happen iteratively until the cost function is minimized and the network is trained. This can be summarized as follows:

**Gradient Descent:** For each layer $l$ and $m$ training samples with learning rate $\eta$,

$$w^l \leftarrow w^l - \frac{\eta}{m}\sum_x \delta^{x,l} \times (a^{x,l-1})^T \tag{9}$$

$$b^l \leftarrow b^l - \frac{\eta}{m}\sum_x \delta^{x,l} \tag{10}$$

Once the parameters of the model are learnt with the aid of the BP algorithm, recognizing an object in an image involves a simple forward propagation of a test image through a sequence of [CONV-POOL] hidden layers to extract the relevant features. Lastly, the feature maps flow through the FC layer which activates certain neurons in this dense network, to recognize the object it is trained for.

Now that we have covered some background on CNNs in this section and on-chip photonics in the previous section, we will describe our proposed *BPLight-CNN* accelerator in the next section.

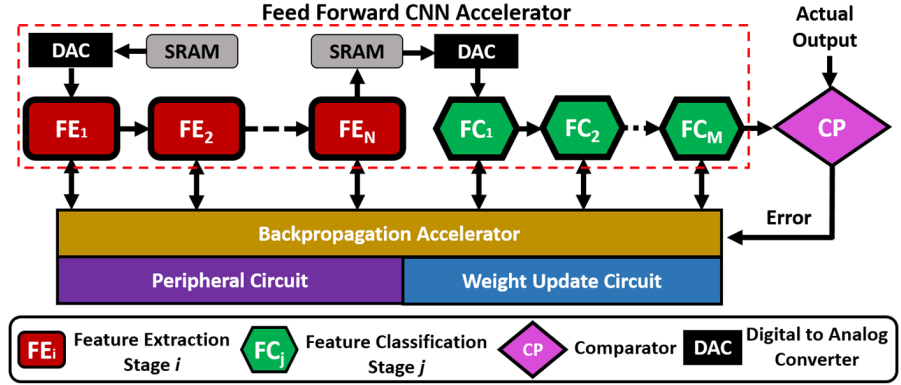

Figure 2: An overview of BPLight-CNN architecture.

## 4. BPLIGHT-CNN ARCHITECTURE
### 4.1 Overview of BPLight-CNN Architecture

Our proposed *BPLight-CNN* architecture is a fully analog, scalable, and configurable memristor-integrated photonic CNN accelerator design. Unlike previously proposed state-of-the-art CNN accelerators [2], [4], *BPLight-CNN* accelerator enables completely analog end-to-end training and testing for a CNN. Fig. 2 gives a high-level overview of this *BPLight-CNN* architecture. As shown in the figure, *BPLight-CNN* comprises of three parts: feedforward CNN accelerator architecture, backpropagation accelerator architecture, and weight update and peripheral circuitry. The proposed analog feedforward CNN accelerator (discussed in subsection 4.2) enables Feature Extraction (FE) through a memristive convolution layer and silicon photonics based ReLU and pooling layers. The feedforward CNN accelerator uses memristive multiplication for Feature Classification (FC). The entire backpropagation accelerator (discussed in subsection 4.3) is implemented in the photonic realm using MRMs, splitters, and multiplexers. Finally, *BPLight-CNN*'s weight update and peripheral circuitry (discussed in subsection 4.4) are implemented through a group of memristors. Furthermore, *BPLight-CNN* architecture scope of work is limited to functioning of resistive-memristors. Analysis on the impact of thermal- and shot-noise on memristors is beyond the scope of this work. Rest of this section describes these three components of *BPLight-CNN* in more detail.

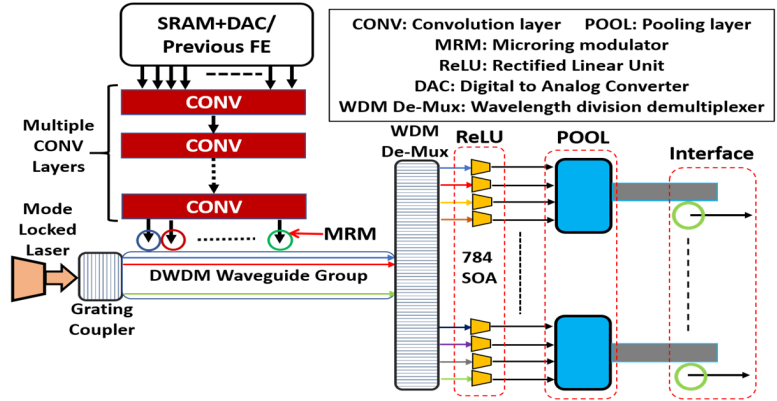

Figure 3: Microarchitecture of Feature Extractor (FE) in BPLight-CNN: showing CONV, ReLU, POOL, and and the Interface layer

## 4.2 Feedforward CNN Architecture

We consider an image dataset as input data and its classification as the application to be executed with *BPLight-CNN*. The CNN accelerator in the proposed *BPLight-CNN* architecture (see Fig. 2) is used for feedforward feature extraction (FE) followed by feature classification of input images. The FE in the CNN architecture is carried out using multiple FE stages ($FE_i$). After all of the features are extracted, feature classification is performed using one or more fully-connected layers (FC). Fig. 3 illustrates the microarchitecture of an FE stage. Each FE stage comprises of multiple memristor-based convolution layers (CONV), a semiconductor-optical amplifier (SOA)-based ReLU layer, an optical comparator based max-pooling (POOL) layer, and an interface layer. *BPLight-CNN's* FE adopts a completely analog computing paradigm by avoiding inter-layer A-to-D (Analog-to-Digital) and D-to-A (Digital-to Analog) conversions compared to state-of-the-art CNN accelerators [2], [4] which use analog memristive convolution and digital CPU/GPU based ReLU and Pooling. The detailed design is discussed in the following subsection.

*4.2.1 CONV Layer Microarchitecture*

CONV is the first step for FE. As shown in Fig. 3, there are multiple CONV layers in each FE. Each CONV layer has multiple weight memristor arrays (WMAs). The basic building block of a WMA is a memristor. A memristor is a metal-oxide-based two-terminal electronic component [22], whose conductance $G$ can be varied by external current flux. A detailed discussion on the operation of a WMA is presented in Section 4.2.1(a). The first CONV layer receives analog data from an SRAM register through a DAC array or from the previous FE stage. An efficient pipelined approach is used to store input data (e.g., image pixels) in the SRAM, and this approach is discussed in Section 5. Intermediate CONV layers receive data from previous CONV layer and transfer data to the next CONV layer. Finally, an array of MRMs receives convolved data from the last CONV layer and modulates that information on carrier wavelengths to transfer this data to the ReLU layer (see Section 4.2.2) for further processing. In addition, FE process employs array of mode-locked lasers to produce light-wave carriers with different wavelengths.

The FE process in the *BPLight-CNN* architecture can convolve 56×56 image pixels in a complete cycle. Furthermore, *BPLight-CNN* processes 56×56 RGB image pixels sequentially in a pipelined manner. The 56×56 input data is divided into 4 chunks of 28×28 input pixels. BPLight-CNN can use multiple chunks (i.e., 8, 16 and so on) of 28×28 input pixels to process input data larger than 56×56, however this analysis is beyond the scope of this work. As explained earlier, before performing convolution, 28×28 input pixels stored in SRAM are converted to analog data using a DAC array. Moreover, an SRAM is connected to the DAC array using eight 128-bit memory buses. To enable conversion of 784 pixels (or 28×28), 13 64-channel DACs are employed. Four WMAs are used are in each CONV layer to process the analog data, where each WMA comprises of 38416 memristors. More information about performing convolution using WMAs is presented in next subsections 4.1.1(a) (i.e., WMA reconfiguration) and IV.A(1)(b) (i.e., memristive convolution). Finally, FE utilizes 196 dense wavelength-division-multiplexing (DWDM) waveguides, each carrying 16 wavelengths. More discussion on the choice of 16 wavelengths is presented in Section 6. This ensures simultaneous traversal of 4 28×28 pixels in a single cycle. In the following sections, the term waveguide group refers to the set of 196 DWDM waveguides.

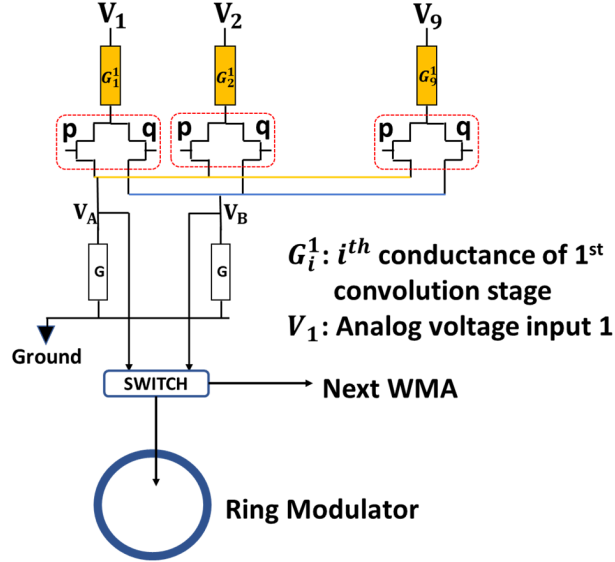

Figure 4: Memristive convolution in a CONV layer; red boxes contain transistor switch for reconfiguration; yellow rectangular boxes are memristors.

*(a) WMA Reconfiguration:* A major parameter to consider during CNN model selection is its filter size. A CNN model can use filters of different sizes such as 1×1, 2×2, 3×3, 4×4, 5×5, and 7×7 etc. The WMA in CONV can be configured to support filters of different sizes. Fig. 4 demonstrates the WMA reconfiguration process to deploy 3×3 filters as an example. To configure a 3×3 memristor filter, memristors in a WMA are divided into multiple 9-memristor based memristor banks. The input of a memristor is either connected to an analog output of a DAC from the DAC-array or is grounded for zero-padding which is required in the convolution. The output terminal of each memristor is connected to two electronic switches p and q. Similarly, to configure a 7×7 memristor bank, all the memristors in a WMA are divided into 784 memristor banks, each comprising of 49 memristors. This novel reconfiguration approach makes the proposed *BPLight-CNN* architecture flexible enough to emulate any CNN model. Another important aspect of convolution is striding (see Section 2.1.1). Analog inputs from the DAC or previous FE stages are connected with each memristor unit considering the required interval during the design time to satisfy striding, e.g. for a 1×1 stride, the WMA ($G_1^C, G_2^C, \ldots G_9^C$) is connected with the 1st set of inputs ($V_1, V_2, \ldots V_9$) and then the next identical WMA ($G_1^C, G_2^C, \ldots G_9^C$) is connected with ($V_2, V_3, \ldots V_{10}$), and so on.

*(b) Memristive Convolution:* Fig. 4 demonstrates memristive convolution using a 3×3 memristor bank as an example. Each memristor bank has 9 memristors with conductance $G_1^C, G_2^C, \ldots G_9^C$. A memristor can be programmed to carry up to 1000 states or conductance values [32]. We chose the value of each conductance $G_i^C$ such that $W_i^C = G_i^C$ (i = 1,2, …9), where $W_1^C, W_2^C, \ldots W_9^C$ are weight elements of a 3×3 kernel as defined in Eq.1 and Eq.2. The kernel elements are chosen randomly in the beginning and then are updated by backpropagation during the training mechanism. The weight update mechanism was explained in Section 2.2.

Convolution with a memristor bank works as follows. Each weight value $W_i^C$ can either be a positive value or be a negative value. As conductance of a memristor cannot be negative, a CMOS switch is used in our design to store negative weight in a memristor. If $W_i^C$ is positive, CMOS switch $p_i$ (see Fig. 4) of the corresponding memristor (i.e., $G_i^C$) is switched ON. If $W_i^C$ is negative, switch $q_i$ (see Fig. 4) is switched ON. Let the analog voltage inputs (i.e., input image pixels converted to analog data) to memristors $G_1^C, G_2^C, \ldots G_9^C$ of the first memristor bank be $V_1, V_2, \ldots V_9$ respectively. In Fig. 4, currents from memristors carrying positive weights are accumulated in terminal $A$ and currents from memristors carrying negative weights are accumulated in terminal $B$ (Kirchhoff's law). The convolved output can be written as $(V_A - V_B)$ where $V_A$ is voltage at A and $V_B$ is voltage at B:

$$C_k = V_A - V_B \qquad (11)$$

where $C_k$ is the resulting voltage or the convolved output from the $k_{th}$ memristor bank, $(V_A = I_p * R)$ and $(V_B = I_q * R)$, $I_p$ is current accumulated from memristors through all switches marked as $p$ and $I_q$ is the current accumulated from memristors through all switches marked as $q$. The current values are:

$$I_p = \sum_{i=1}^{9} [V_{9*k+i} * G_i^C] \quad \text{for } W_i^C > 0 \qquad (12)$$

$$I_q = \sum_{j=1}^{9} [V_{9*k+j} * G_j^C] \quad \text{for } W_j^C < 0 \qquad (13)$$

Each convolved output $C_k$ is fed to the peripheral circuit as it will be used by the backpropagation architecture later. Details of the peripheral circuit are discussed in Section 4.4. Apart from the peripheral circuit, each $C_k$ is input to a microring modulator (MRM) for data modulation. There are 784 MRMs each of which can modulate a light-wave in the DWDM waveguide. A modulated photonic signal with wavelength $\lambda_k$ in a photonic waveguide can be expressed as:

$$L_k = C_k * A \sin(\frac{2\pi}{\lambda_k} t + \theta) \qquad (14)$$

where $L_k$ is modulated light-wave with wavelength $\lambda_k$, carrying convolved data $C_k$, A is the amplitude of the $k_{th}$ light-wave before the data modulation phase. By setting A = 1,

$$L_k = C_k \sin(\frac{2\pi}{\lambda_k} t + \theta) \qquad (15)$$

After data modulation, all the light-waves $L_k$ (k = 1, 2, ..., 784) are decoupled from the DWDM waveguide by a DWDM de-coupler. After decoupling, each individual light-wave is fed to a semiconductor-optical-amplifier (SOA) in the ReLU layer.

*4.2.2 ReLU Layer Microarchitecture*

As discussed in Section 3, an SOA is a silicon photonic component used to amplify a photonic signal. An SOA uses an electronic pumping mechanism to provide gain to an input photonic signal. The electronic pump current to an SOA can be varied to set its total gain. The characteristics are almost linear when an SOA's gain is close to 1. This linear behavior is identical to ReLU (see Eq. (3)) which is a widely used deep learning neural activation function. In addition to simple linear amplification, it has been demonstrated in [8] that an SOA can be tuned to emulate other neuron functions that are used in deep learning, such as Tanh, Sigmoid, exponential, etc, which broadens the scope of our *BPLight-CNN* accelerator neurons. Therefore, we set the gain of all the SOAs in our design to 1. There are 784 SOAs in a ReLU layer of *BPLight-CNN* as shown in Fig. 3. The $k_{th}$ SOA takes light-wave $L_k$ as input and produces the following output.

$$ReLU_k = \begin{cases} 0, & C_k \leq 0 \\ L_k, & C_k > 0 \end{cases} \qquad (16)$$

This is according to the ReLU model explained in Eq. 3. The outputs $ReLU_k$ of all SOAs are subsequently fed to the max-pooling layer, which is discussed next.

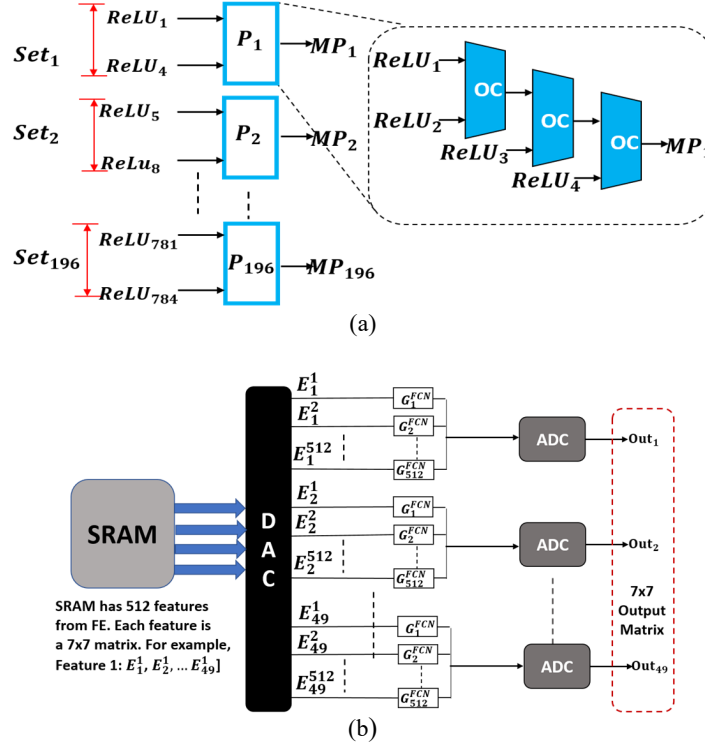

Figure 5: (a) Cascaded optical comparator in POOL, (b) Fully Connected Layer (FC).

*4.2.3 POOL Layer Microarchitecture*

The VGG [42] and LeNet [43] benchmarks (see Table 1) that are used in this work operate on a 2×2 max-pooling with a stride of 2. Therefore, for max-pooling we consider a 2×2 window with a stride of 2. To facilitate 2×2 max-pooling for 784 outputs from the ReLU layer, the photonic max-pooling layer (POOL) uses 196 4-input max-pooling units such as $P_1, P_2, P_3, ... P_{196}$. Each max-pooling unit consists of a cascaded optical comparator arrangement to perform max-pooling. As shown in Fig. 5(a), three 2-channel optical comparators are cascaded to form a 4-input max-pooling unit. For benchmarks operating on higher order max-pool (e.g. 3×3), proportional number of optical comparators can be integrated to design the desired max-pooling unit. We consider a high-speed two-channel optical comparator identical to [20] designed for a nominal wavelength of 1560 nm. The 784 outputs from the ReLU layer are bundled into 196 sets and each set $j$ is fed to max-pooling unit $P_j$. Assuming $ReLU_1, ReLU_2, ReLU_3, ReLU_4$ all belong to set 1 and are input to $P_1$, the max-pooling output of unit $P_j$ can be written as:

$$MP_j = max\{ReLU_{4(j-1)+1}, ReLU_{4(j-1)+2}, ReLU_{4(j-1)+3}, ReLU_{4(j-1)+4}\} \quad (17)$$

The output from the POOL layer is directed to interface layer which has a series of MRM detectors. These detector MRMs perform optical to electrical (O/E) conversion of data before feeding it to the next convolution layer or FC layer.

*4.2.4 FC Layer Microarchitecture*

After feature extraction is performed using the FE stages (by using the CONV, ReLU, and POOL microarchitectures discussed in the previous subsections), features are sent to a feature classification phase. In CNN, the feature classification segment can be viewed as a special case of convolution, where each extracted feature map uses the largest possible kernel. In other words, feature classification comprises of one or more fully-connected (FC) layers. All the layers discussed in the previous subsections are used for feature extraction, whereas the FC layer performs feature classification.

*BPLight-CNN* employs a memristive matrix vector multiplication (M-MVM) based FC layer. The working principle of M-MVM based FC is similar to that of memristive convolution. Fig. 5(b) illustrates a logical layout of a M-MVM based FC layer. When all the features from the feature extraction stage are stored and available in the SRAM buffer, the features are fed to the DAC array of the FC layer as depicted in the figure. As an example, we consider 512 features coming from the feature extraction (FE) stages. VGG and LeNet operate on a 7×7 kernel in FC. Therefore, each feature is a 7×7 matrix, e.g. the i$^{th}$ feature is $E_1^i, E_2^i, \ldots E_{49}^i$. FC has 49 identical memristor banks each of which has 512 memristors: $G_1^{FC}, G_2^{FC}, \ldots G_{512}^{FC}$. Each memristor $G_i^{FC}$ represents an FC weight $W_i^{FC}$ which is obtained through offline training. After analog conversion, each analog value of a feature is applied as voltage across a memristor of the FC's memristor bank. For example, a voltage of $E_1^i$ is applied across $G_1^{FC}$, voltage of $E_2^i$ across $G_2^{FC}$, etc. Depending on whether the weight $W_i^{FC}$ corresponding to memristor $G_i^{FC}$ is positive or negative, the p or q switch is set respectively, (see Fig.4). The accumulated current from each memristor bank is fed to the next FC stage until the final FC stage is reached. Also, outputs from each FC stage are fed to the peripheral circuit to be used by the backpropagation architecture for weight update. The outputs from the final FC stage are the classified outputs for a feedforward CNN. During training, the classified outputs and target outputs are input to an analog subtraction unit, the result of which is fed to the backpropagation architecture, as discussed next.

### 4.3 Backpropagation Architecture

*BPLight-CNN*'s backpropagation (BP) architecture employs analog microring modulators, photodiodes, multiplexers, and splitters to perform completely analog matrix-multiplication and other arithmetic operations. In contrast, previously proposed CNN accelerators [2], [4] adopt a hybrid approach by using analog memristors for matrix multiplications and digital CPU/GPU for other arithmetic operations, which requires performance hindering A-to-D and D-to-A conversions.

Our analog BP architecture mainly involves computing matrix-vector multiplication in the backward pass. A photonic modulator is used for analog amplitude modulation of a light carrier. In its simplest term, analog amplitude modulation is the multiplication of a scalar input with an analog signal. The authors in [18], [39] have demonstrated photonic modulator based analog multipliers. Fig. 6 illustrates the microarchitecture of the proposed BP accelerator design. It is based on photonic matrix-vector multiplication using MRMs (which were discussed in Section 3). We use MRMs for their high accuracy and quality factor [23].

We now describe the operation of the proposed BP architecture. As discussed in Eq. (7), the error at the final layer (*l=L*) of BP is $\delta^{x,L} \leftarrow \nabla_a C_x \odot \sigma'(z^{x,L})$. Here, $\nabla_a C_x$ is rate of change of output w.r.t the output activation (i.e., difference of actual classified output from FC of CNN architecture and the target output). $\sigma'(z^{x,L})$ is the derivative of the ReLU function in the final FC stage of the CNN architecture. Outputs from the final FC stage of the CNN architecture are fed to an analog subtraction and multiplication unit to determine $\delta^{x,L}$. Using Eq. (8) and the computed $\delta^{x,L}$, we calculate error for the *(L-1)$^{th}$* layer using the following equation:

$$\delta^{x,L-1} \leftarrow ((w^L)^T \times \delta^{x,L}) \odot \sigma'(z^{x,L-1}) \quad (18)$$

where, $w^L$ is weight matrix obtained from $L^{th}$ layer of CNN architecture through the peripheral circuit. The details of the peripheral circuit are explained in the next subsection.

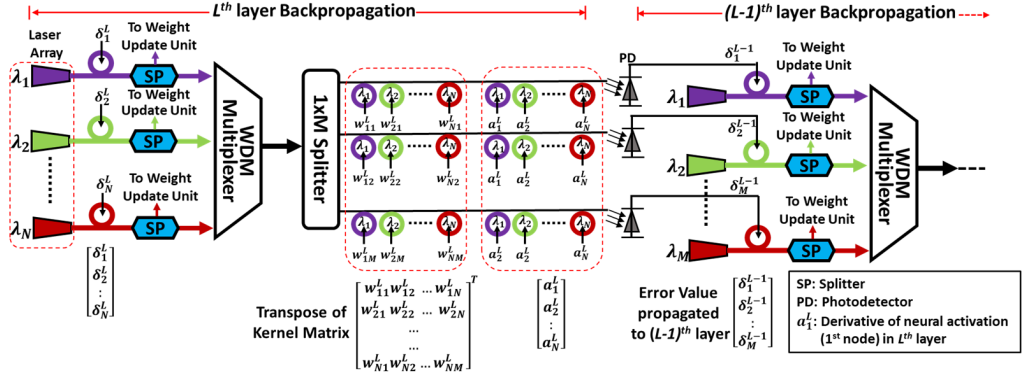

Figure 6: Backpropagation architecture in *BPLight-CNN* which presents the backpropagation between the final layer l=L and penultimate layer l=L-1.

Fig. 6 shows the backpropagation between the final layer *l=L* and its penultimate layer *l=L-1*. As illustrated in Fig. 6, there are N number of wavelength carriers coming from a mode-locked laser array. The value of N for a layer equals to the output feature size for the corresponding layer in the CNN architecture, e.g. N equals 49 (7×7) for the last layer. Each wavelength in layer *L* is modulated with error $\delta^{x,L}$ by an MRM tuned to that wavelength. In Fig. 6, the violet MRM is tuned to modulate $\lambda_1$. Now the *j*th MRM's (where $1 \leq j \leq N$) modulation output on $\lambda_j$ is $MRM_j = \delta_j^{x,L} * A \sin(\frac{2\pi}{\lambda_j} t + \theta)$. Each $MRM_j$ is split into two equal parts. The first part is sent to the weight-update circuitry (see Section 4.4) to update the corresponding weights in the CNN architecture. The other part is fed to a WDM multiplexer. A WDM multiplexer is used to combine multiple light wavelengths into a single multi-wavelength carrier. After multiplexing, the combined optical signal is split into *M* parts where *M* equals the number of neurons (i.e, $1 \leq m \leq M$) in layer *L-1*. Each part is fed to a multi-wavelength waveguide. As a result, in each waveguide there are N wavelengths each carrying data $\delta_{j,n}^{x,L} * B \sin(\frac{2\pi}{\lambda_j} t + \theta)$, where $1 \leq n \leq N, B = \frac{A}{2N}$. Each weight $w_{ij}^L$ of the transpose of $w^L$ obtained from the peripheral circuit is modulated to a light carrier. This results in:

$$M_{i,n} = w_{im}^L * \delta_{j,n}^{x,L} * A \sin(\frac{2\pi}{\lambda_j} t + \theta) \qquad (19)$$

Now, each $M_{i,n}$ is modulated with $a_n^L$ which is a derivative of the ReLU functions of layer *L-1* (equal to $\sigma'(z^{x,L-1})$ in Eq. (18)). Then, $M'_{i,n}$ becomes,

$$M'_{i,n} = w_{im}^L * \delta_{j,n}^{x,L} * a_n^L * A \sin(\frac{2\pi}{\lambda_j} t + \theta) \qquad (20)$$

Next, a photodiode is used to demodulate photonic data from each waveguide. The photodiode demodulates the combined output $M'_{i,n}$ for all wavelengths in a waveguide which is nothing but the matrix-vector multiplication identical to Eq. (18). The output of each photodiode is passed through a signal conditioning and filtering circuit to remove unwanted noises. Details of the conditioning circuit are omitted for brevity. The output from the signal conditioning circuit looks as follows:

$$\delta^{x,L-1} = ((w^L)^T \times \delta^{x,L}) \odot a^L \qquad (21)$$

where, $\delta^{x,L-1}$ is the error to be propagated from layer *(L-1) to (L-2)*. The same procedure as above is continued until the 1st layer is reached. While doing the backpropagation, the error value in each layer is also fed to the corresponding weight-update circuit, which is discussed in more detail below.

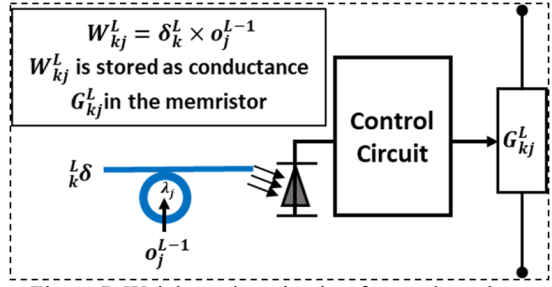

Figure 7: Weight-update circuitry for any layer l

### 4.4 Weight Update and Peripheral Circuitry

*4.4.1 Weight-update circuitry*

For weight-update, each element of a weight kernel in any layer $l$ of CNN architecture can be written as $w_{k,j}^l$. Please note that $l$=L for the final layer. Each $w_{k,j}^l$ is stored in a memristor of a memristor bank in layer $l$ as $G_{k,j}^l$ (as explained in Section 4.2.1). The weight-update equation for $w_{k,j}^l$ (or, $G_{k,j}^l$) can be written as per Eq. (9), as follows:

$$G_{new(k,j)}^l \leftarrow G_{old(k,j)}^l - \frac{\eta}{m} \times \delta_k^l \times O_j^{l-1} \qquad (22)$$

where, $O_j^{l-1}$ is the $j^{th}$ output from the POOL of the *(l-1)* layer of the CNN architecture. Fig.7 illustrates the weight-update circuitry for any layer $l$. As shown in Fig. 7, $\delta_k^l$ is obtained from the BP architecture as a photonic signal. $O_j^{l-1}$, which is collected from the peripheral circuit, is used to modulate the light carrier carrying the error value $\delta_k^l$. The modulated output is demodulated using a photodiode and then sent to a signal conditioning circuit. In the signal conditioning circuit, first the analog signal is filtered (from noises) and passed through a subtractor to obtain new $G_{k,j}^l$ as depicted in Eq. (20). The previous conductance or weight value $G_{old(k,j)}^l$ is fed to the subtractor from the $l^{th}$ layer memristor bank. The new conductance value $G_{k,j}^l$ is now fed to the equivalent memristor control circuit to update its weight value. The conditioning circuit as well as the memristor control circuit are from [3].

*4.4.2 Peripheral Circuitry*

The output $MP_j$ from the POOL of a layer $l$ can be written as $MP_j^l$. During the feedforward training phase, each $MP_j^l$ is stored as conductance in a memristor in the peripheral circuitry. This is used in backpropagation as $O_j^l$, an output of the $l^{th}$ layer (as per Eq. (22)). Each $MP_j^l$ is sent to a signal conditioning circuit and then a memristor control circuit. The resulting electronic signal is used to update the conductance (or weight value) of the memristor.

### 5. BPLIGHT-CNN CASE STUDY

In this section, we demonstrate the working principle of a pipelined *BPLight-CNN* architecture for a CNN benchmark VGG [42] on the ImageNet dataset [45]. We select a particular configuration, namely, VGG-A for the case study. However, we also experiment with all variants of the VGG [42] and LeNet [43] benchmarks as shown in Table 1 and discussed in Section 6. Using microarchitectures of the convolution layer, ReLU layer, POOL layer, interface layer, and FC layer as explained in Section 4, we configured *BPLight-CNN* as illustrated in Fig. 8(a) for VGG-A application with four FE stages. The details of it are as follows.

VGG for the ImageNet dataset operates on a 224×224 image input. As explained in Section 4.2.1, *BPLight-CNN* can convolve 56×56 pixels at a time, i.e., one *BPLight-CNN* cycle. Therefore, it requires 16 *BPLight-CNN* cycles to execute a 224×224 image. Please note that a *BPLight-CNN* cycle is different from its clock cycle. Here, one *BPLight-CNN* cycle refers to the complete feature extraction and feature classification of a 56×56 image. The SRAM register array in *BPLight-CNN* is of size 2 KB to store the 56×56 input data. CONV performs feature extraction on a 28×28 input data at a time in a pipelined manner. FE in *BPLight-CNN* is performed as explained in the CONV architecture (Section 4.4.1).

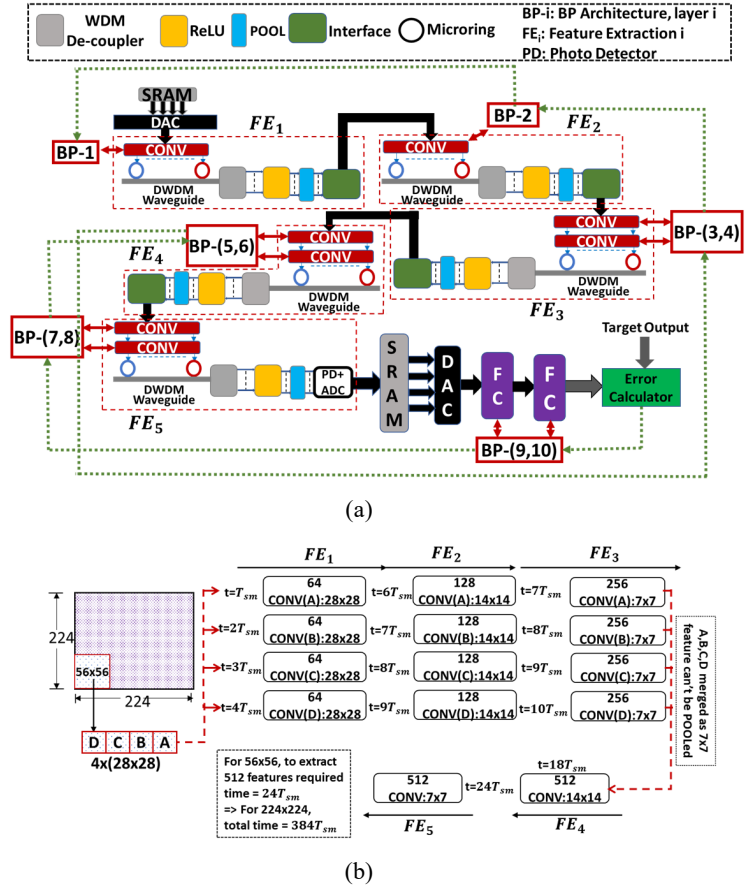

Figure 8: (a) VGG-A implemented on *BPLight-CNN* (b) Pipelined dataflow in feedforward operation in *BPLight-CNN*.

Fig. 8(b) demonstrates the pipelined dataflow of the feedforward operation in *BPLight-CNN*. We consider a 2.5 GHz clock. Therefore, the clock cycle period $T_{sm}$ = 400 ps. As shown in Fig. 8(b), at t=$T_{sm}$, the first set of 28×28 pixels from SRAM (i.e., A) are convolved (64 filters/features) and are stored in memristors in the peripheral circuit. The other three set of 28×28 pixels are namely, B, C, and D. Note that CONV convolves a 28×28 input in one clock cycle. As $FE_1$ for VGG-A consists of one convolution layer (see Table 2), convolved outputs of CONV-1 of $FE_1$ is directly sent to the modulation phase. In the modulation phase, each convolved output is modulated by an MRM of a particular tuning wavelength to a light carrier of that wavelength in the DWDM waveguide group. The DWDM waveguide group (i.e., as mentioned in Section 4.2.1 each waveguide group has 196 waveguides with 16 DWDM wavelengths in each of the waveguide) can accommodate 4×784 wavelengths or in other words 4 features of size 28×28. The time required for convolved data of one FE to arrive at the next FE, $T_{FE}$ = modulation time + ReLU time + POOL time + interface time = 20 ps + 10 ps + 10 ps + 10 ps = 50 ps. From $t=T_{sm}$ to $t=2T_{sm}$, CONV(A) outputs from the peripheral circuit of $FE_1$ are modulated, ReLU and POOL'ed, and then fed to $FE_2$. There can be 8 such data movements as $\frac{T_{sm}}{T_{FE}} = 8$. In one data movement, 4 28×28 features can be processed. Therefore, at $t=2T_{sm}$, 32 CONV(A) features arrive at $FE_2$. Similar to CONV(A), from $t=2T_{sm}$ to $t=3T_{sm}$, 32 CONV(B) features; from $t=3T_{sm}$ to $t=4T_{sm}$, 32 CONV(C) features; from $t=4T_{sm}$ to $t=5T_{sm}$, 32 CONV(D) features are convolved and stored in the peripheral circuit of $FE_2$. After this, from $t=5T_{sm}$ to $t=6T_{sm}$, the remaining 32 CONV(A) features in $FE_1$ are convolved in $FE_2$. In this way, by $t=6T_{sm}$, all the 64 CONV(A) features in $FE_1$ are convolved with 128 $FE_2$ filters to produce 128 features and stored in the memristors of its peripheral circuit. Similarly, remaining 32 B, C, and D features are convolved and stored (Fig. 8(b)) by $t=7T_{sm}$, $t=8T_{sm}$, and $t=9T_{sm}$ respectively. $FE_1$ has 64 features, $FE_2$ has 128 features, $FE_3$ has 256 features, etc, as per the VGG-A configuration (Table 1). It is important to note that 64 CONV(A) features from $FE_1$ are convolved with 128 memristive WMAs (kernels/filters) to produce 128 CONV(A) features for $FE_2$. Similarly, 128 CONV(A) features from $FE_2$ are convolved with 256 WMAs to produce 256 CONV(A) features for $FE_3$.

Table 1. CNN Benchmark Configuration For VGG, LeNeT

|  | $FE_1$ | $FE_2$ | $FE_3$ | $FE_4$ | $FE_5$ |  |
|---|---|---|---|---|---|---|
| VGG-A | 3×3, 64, 1 | 3×3, 128, 1 | 3×3, 256, 2 | 3×3, 512, 2 | 3×3, 512, 2 | FC-4096,2 FC-1000, 1 |
| VGG-B | 3×3, 64, 2 | 3×3, 128, 2 | 3×3, 256, 2<br>1×1, 256, 1 | 3×3, 512, 2<br>1×1, 256, 1 | 3×3, 512, 2<br>1×1, 256, 1 |  |
| VGG-C | 3×3, 64, 2 | 3×3, 128, 2 | 3×3, 256, 3 | 3×3, 512, 3 | 3×3, 512, 3 |  |
| VGG-D | 3×3, 64, 2 | 3×3, 128, 2 | 3×3, 256, 4 | 3×3, 512, 4 | 3×3, 512, 4 |  |
| LeNET-A | 3×3, 6,1 | 3×3, 6,1 | 3×3, 16,2 | 3×3, 16, 4 | 3×3, 120, 1 | FC8 4,1 |
| LeNET-B | 3×3, 6,1 | 3×3, 6,1 | 3×3, 256, 1 | 3×3, 16,6 | 3×3, 120, 1 |  |

A, B, C, and D are convolved separately until $t = 10T_{sm}$ when all of them arrive at $FE_3$ as 256 7×7 features each. Now, all of these features are merged together to form 256 28×28 features. Therefore, it will require another $8T_{sm}$ time (i.e., $t=10T_{sm}$ to $t=18T_{sm}$) to send 256 28×28 features from $FE_3$ and convolve them as 512 14×14 features at $FE_4$. Similarly, convolution, ReLU, and POOL are performed in $FE_4$ and $FE_5$. As illustrated in Fig. 8(b), at $t=24T_{sm}$, 512 features are obtained from $FE_5$ for 56×56 pixels. As shown in Fig. 8(a), features from $FE_5$ are stored in SRAM until all the 224×224 pixels are extracted. For 224×224 pixels, it will take 16×24$T_{sm}$=384$T_{sm}$=153.6ns. After this, all the features are retrieved from SRAM and fed to FC for feature classification. The first FC operation requires $(T_{sm} + T)$ time as it is identical to FE. The second FC operation requires $T$ time as no more SRAM read is needed. This means that *BPLight-CNN* requires 153.6 ns (for FE) +$T_{sm} + 2T$ = 154 ns, for one forward pass.

After a forward pass, the FC output is sent to the BP architecture for backpropagation. Each layer in BP requires $T_b$ units of time where $T_b$ = (error modulation to light carrier along with its driver time) + (split time) + (WDM multiplexing time) + (split time) + (weight modulation time) + (ReLU function derivative modulation time) + (demodulation time along with photodiode and transimpedance amplifier time) = 20 ps + 10 ps + 10 ps + 10 ps + 10 ps + 10 ps +20 ps = 90 ps. It takes $6T_b$ units of time to complete one backward pass.

In summary, *BPLight-CNN* requires 154 ns for one forward pass and 90 ps for a backward pass. The ultra-fast nature of photonic interconnects allows for high-speed backpropagation in *BPLight-CNN*.

## 6. EXPERIMENTAL ANALYSIS
### 6.1 CAD for BPLight-CNN

We use IPKISS [46], a commercial optoelectronic CAD tool, to design and synthesize all of the photonic components of *BPLight-CNN*. All of the synthesized components are integrated together to design *BPLight-CNN*. For all of the photonics components, we consider a 32nm IPKISS library, similar to prior work [47]. The parametric details for *BPLight-CNN* are shown in Table 2. We developed a C++ based architectural simulator which takes device- and link-level parameters from IPKISS, to estimate performance of *BPLight-CNN* accelerator for several benchmarks.

*6.1.1 Power and Area Models*

The power and area of all *BPLight-CNN* components are summarized in Table 2. A DWDM wavelength range in the C and L bands [25], with a starting wavelength of 1550 nm [26] is considered for our analysis. MRs in *BPLight-CNN* are expected to be in partial resonance with their corresponding wavelengths in the waveguide. This MR partial resonance was modeled based on the circuit-level analysis presented in prior work [30], [48], [50]. In addition, we consider an MR heating power of 15 µW [31] per MR. Note that the power overheads of mitigating process variations are not considered and are beyond the scope of this work. Furthermore, we use Caphe [46] for modeling power and area of all photonic elements such as modulators, demodulators, waveguides, lasers, etc. The energy and area parameters for memristors are adapted from [2]. We adapted power and area models for DAC from [41]. We also use power and area parameters from [3] for the ADC array used in the FC layer of *BPLight-CNN*.

*6.1.2 Performance Models*

We use Caffe [38], a deep learning framework, to train the datasets in conjunction with photonic component results from IPKISS. We program the switches in each CONV layer to map each of our benchmarks in BPLight-CNN. This ensures zero pipeline hazards between any two layers in BPLight-CNN. Currently both VGG and LeNET require only a few CONV layers therefore we choose to program their corresponding switches manually. We can easily automate this process for a workload which requires tens to hundreds of CONV layers by storing each switch configuration as a bit in an SRAM which will be connected to all CONV layer switches to program them. We determine computational efficiency, energy efficiency, throughput, and prediction error rate to compare the performance of *BPLight-CNN* with a state-of-the-art CNN accelerator, namely PipeLayer [4]. We also use GPU results (from [4]) as the baseline for comparison. We evaluate for the following metrics: *Computational efficiency* represents the total number of fixed point operations performed per unit area in one second (GOPS/s/mm$^2$); *Energy efficiency* refers to the number of fixed point operations performed per watt (Giga operations per watt or GOPS/s/W); *Throughput* is the total number of operations per unit time (GOPS/s); and lastly, *Prediction error rate* is the percentage of error in inferring any datasets.

*6.1.3 Benchmarks and Datasets*

We use two widely used CNN benchmarks: VGG-Net and LeNet [45]. We consider four variants of the VGG benchmark: VGG-A, VGG-B, VGG-C, and VGG-D and two variations of LeNet (LeNet-A and LeNet-B). The configuration of all stages of VGG and LeNet benchmarks for these variants are depicted in Table 1. In the table, CONV-I represents convolution stage 'I' for a benchmark model. "M×M, K, N" for a convolution stage means that the convolution stage comprises of M×M filters, and N number of back-to-back convolution layers, with each convolution layer having convolutional width K. The convolutional width is the number of convolutional filters in a convolution layer. Furthermore, we do consider a unit size window stride for the benchmark variants. For VGG, we use ImageNet dataset [45] having 224×224 images for training and inference. For LeNet, we use 28×28 images of MNIST datasets [21] for training and inference.

TABLE 2. *BPLIGHT-CNN* Parameter Details [4], [31], [35]-[37], [49]-[51]

| Components | Parameters | Values | Power (mW) | Area (mm²) |
|---|---|---|---|---|
| SRAM register | Size | 2KB | 10 | 0.2 |
| | Count | 16 | | |
| DAC | Resolution | 16-bit | 4.374 | 0.000208 |
| | Frequency | 1.2 Gbps | | |
| | Channel | 64 | | |
| | Count | 208 | | |
| ADC | Resolution | 16-bit | 490 | 0.294 |
| | Frequency | 1.2 Gbps | | |
| | Count | 245 | | |
| WRA | Number | 48 | 24.5 | 0.000514 |
| Memristor banks (512 per bank in FCN) | Number | 49 | 0.45 | 0.000003 |
| Modulator | Time | 20ps | 1080.8 | 39.38 |
| | Count | 62720 | | |
| De-Modulator | Time | 20ps | 1080.8 | 39.38 |
| | Count | 62720 | | |
| Trans-Impedance-Amplifiers (TIA) | Time | 10ps | 0.18 pJ/bit | 0.28 |
| | Count | 62720 | | |
| Electrical Comparator | Time | 180 ps | 0.02 | 0.00049 |
| | Count | 784 | | |
| WDM coupler | Count | 16 | 0 | 0.00028 |
| WDM de coupler | Count | 16 | 0 | 0.00028 |
| Optical comparator | Time | 60 | 0 | 0.0045 |
| | Count | 980 | 0 | |
| Mode-locked laser | Wavelengths | 16 | 32000 | 0.384 |
| | Count | 6 | | |
| Waveguide | DWDM | 16 | 0 | 80 |
| | Width | 450 nm | | |
| | Count | 520 | | |

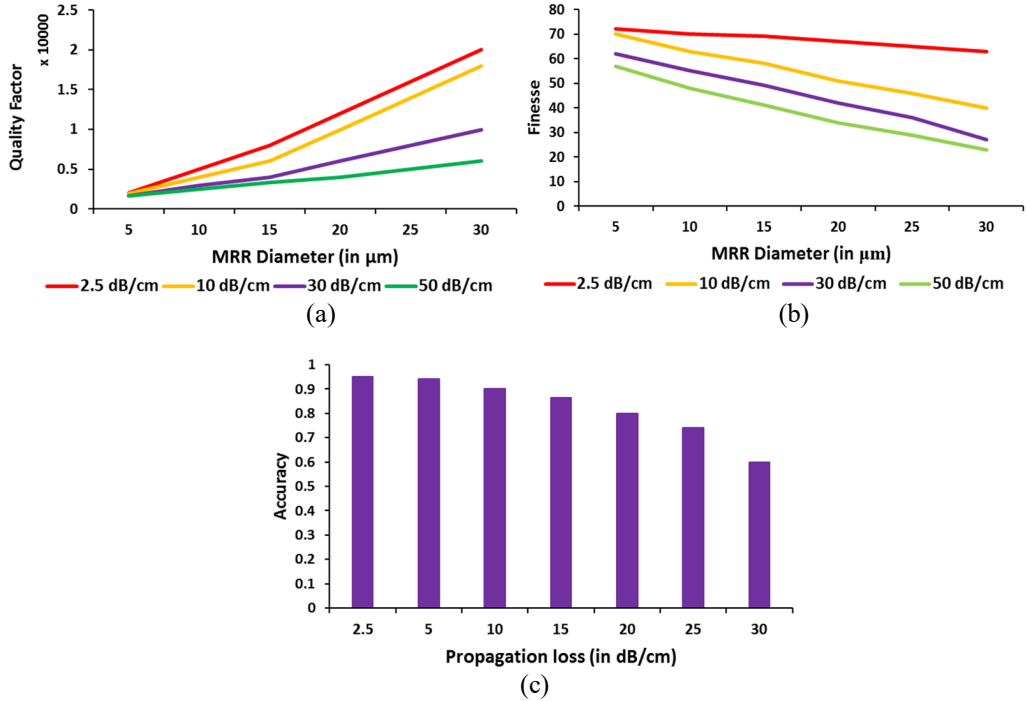

Figure 9: (a) MRM Q-factor (b) MRM Finesse (c) average prediction accuracy w.r.t propagation loss in photonic components diameter (assuming a 32-bit weight resolution).

## 6.2 Sensitivity Analysis with Prediction Accuracy

MRMs are used extensively in the *BPLight-CNN* design, both in the feedforward and BP architectures. The prediction accuracy of *BPLight-CNN* depends on losses encountered by the photonic signal when it traverses through the photonic waveguide, MRMs, and other photonic components. These losses degrade photonic signal intensity before it reaches SOA (which acts as ReLU in *BPLight-CNN*), and causes the SOA to operate in a non-linear region, reducing the overall prediction accuracy. Among all of the losses, the MRM's insertion loss and waveguide propagation loss are the major contributors to prediction error in *BPLight-CNN*. The MRM's insertion loss depends on its Quality factor (Q-factor) and finesse. Q-factor is the number of photonic cycles taken by a photonic signal before its intensity goes to zero in an MRM. Finesse is the number of photonic cycles before a photonic signal's intensity becomes 70.7% of its initial value. As both Q-factor and finesse are determined by the MRM diameter, therefore, in this section we present a sensitivity analysis to determine the optimal MRM diameter.

Fig. 9(a) and 9(b) illustrate the MRM's Q-factor and finesse w.r.t. its diameter (in μm), respectively. From Fig. 9(a), it can be seen that increase in MRM diameter leads to higher Q-factor, which ultimately leads to lower insertion loss. On the other hand, from 9(b) it can be observed that increase in MRM diameter decreases its finesse, and increases insertion loss. Therefore, we select an optimal MRM of diameter of 10μm to minimize overall insertion loss. Furthermore, we have considered an FSR of 20nm based on the device-level analysis presented in prior work [34] on an MRM of diameter 10μm. In line with this MRM FSR, each waveguide is expected to have 16 DWDM with a channel spacing of 1.25nm (see Table 2). Considering this MRM diameter, Fig. 9(c) presents average prediction accuracy variation with increase in waveguide propagation loss (in dB/cm) for all the applications discussed in Section 6.1.3. In this analysis, we have considered photonic waveguide groups of fixed lengths in different parts of the *BPLight-CNN* architecture, where each waveguide in a waveguide group is coupled with a fixed number of MRMs with 10μm diameter. From this plot it can be seen that increase in photonic waveguide propagation loss decreases prediction accuracy. An increase in waveguide propagation loss decreases photonic signal integrity and decreases predication accuracy. In addition, increased waveguide propagation loss also increases insertion losses of MRMs which increases overall losses and worsens prediction accuracy further. Therefore, we have considered the waveguide propagation loss of 2.5 dB/cm [24] for the rest of our analysis. The worst-case power loss path in our BPLight_CNN architecture will be in our Feedforward architecture, as the Backpropagation architecture performs optical to electrical conversions between its layers. Therefore, the worst-case power loss path will be from the grating coupler to ReLU (i.e., SOA) of the Feature Extractor (FE) process shown in Fig. 3. Furthermore, we used SOA sensitivity of -20dBm [27], MR through loss of 0.02 dB [28], and waveguide coupler/splitter loss of 0.5dB [28] to calculate the worst-case power loss of BPLight-CNN, to determine the photonic laser power and correspondingly the electrical laser power considering wall-plug laser efficiency of 3% [29]. Finally, the SOA sensitivity is sufficient to mitigate crosstalk noise effects and yield a bit-error-rate (BER) of 10-9. Therefore, crosstalk noise in BPLight-CNN has negligible impact on optical data computation accuracy.

There are other minor factors which affect the prediction accuracy of *BPLight-CNN*: (1) Each memristor can have 1000 quantized states. The quantization error encountered due to limited number of memristor states contributes up to 1.2% of Prediction Error (PER); (2) The signal-to-noise ratio of SOA used in *BPLight-CNN* is 50 dB, which is adapted from [8]. The SOA's contribution to the overall PER is 2.35%; (3) Each optical comparator in *BPLight-CNN* has an SNR of 40 dB [42]. This accounts for a PER of 1%; and (4) the memristor-photonic interface is noisy. The signals from memristors going to modulators encounter a noise with an SNR of 25 dB which leads to a PER of 1.45%. We obtained these numbers through detailed optoelectronic synthesis using the IPKISS tool.

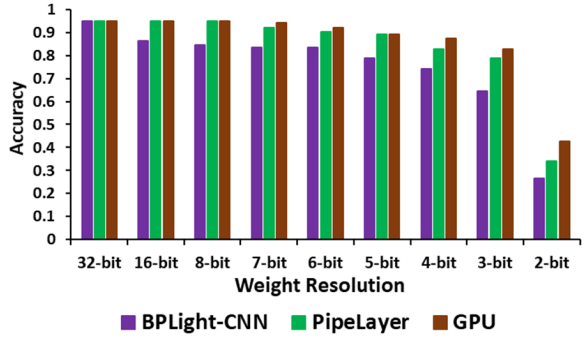

Figure 10: *BPLight-CNN* average prediction accuracy comparison with PipeLayer [4] and GPU-based execution across different weight resolutions varying from 2-bit to 32-bit.

We conducted another sensitivity analysis to explore the impact of weight resolution on average prediction accuracy. Fig. 10 compares the accuracy of the proposed *BPLight-CNN* with PipeLayer [4] and GPU-based execution across different weight resolutions from 2-bit to 32-bit. From this plot it can be seen that the accuracy of *BPLight-CNN* increases with increase in weight resolution, due to the resulting reduction in quantization error across *BPLight-CNN*. Interestingly, *BPLight-CNN* achieves a prediction accuracy of 95% (i.e., slightly lower than state-of-the-art GPU accuracy of 96% and PipeLayer accuracy of 95.6%) when its weight resolution is 32-bit. The DACs used in *BPLight-CNN* can be configured upto a precision of 16-bit. Therefore, we use a 16-bit weight resolution in our performance and energy analysis.

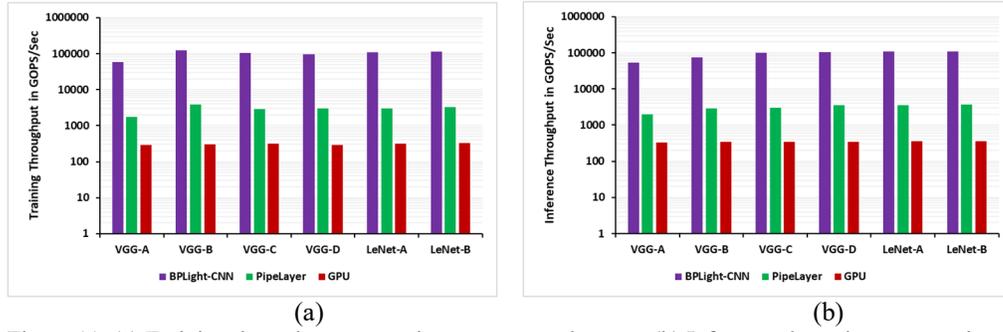

(a)                       (b)

Figure 11: (a) Training throughput comparison across accelerators, (b) Inference throughput comparison across accelerators.

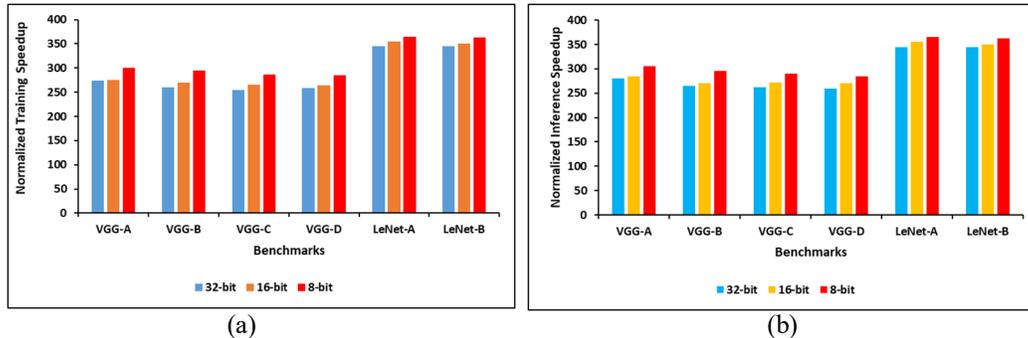

(a)                       (b)

Figure 12: Normalized Speedup of BPLight-CNN w.r.t weight resolution, (a) for training, and (b) for inference. Here the training and inference speedup are normalized w.r.t GPU for different weight resolutions.

## 6.3 Performance analysis

Fig.11a and Fig.11b demonstrate training and inference throughputs (respectively) of *BPLight-CNN* and PipeLayer [4], and a baseline GPU implementation [4], for four variations of the VGG and two variants of the LeNet benchmarks. The results are almost identical for both training and inference. The GPU-based accelerator performs with an average throughput of 305 GOPS/s during training and 347 GOPS/s during inference. PipeLayer shows an average throughput of 2923 GOPS/s during training and 3102 GOPS/s during inference. The proposed *BPLight-CNN* shows an average throughput of 99534 GOPS/s during training and 90985 GOPS/s during inference. The superior performance of *BPLight-CNN* is due to the intelligent integration of ultra-fast memristors and high-speed photonic components such as MRAs, SOAs, and comparators. The overall throughput of PipeLayer is affected by inter-layer data conversion with relatively slow ADCs. Also, PipeLayer spends most of its time in sequential weight updates during training. However, *BPLight-CNN* has an inherent advantage due to its photonic parallel weight update mechanism. On average, *BPLight-CNN* outperforms PipeLayer by 34× for training and 29× during inference; and the proposed accelerator outperforms GPU by 326× for training and 263× for inference. Finally, for the results presented in Fig. 11, the variance of speedup across benchmarks is 1650 with a standard deviation of 40.02.

Fig. 12a and 12b illustrate the effects of weight resolution on training and inference speedups of *BPLight-CNN* respectively. For both training and inference, with the rise in weight resolution, there is a gradual degradation in speedup. The 32-bit speedup is 1.5% lower compared to 16-bit and 8% lower compared to 8-bit during training. During inference, the 32-bit speedup is 3.5% lower compared to 16-bit speedup and 11% lower compared to 8-bit speedup. This is due to the additional delay in storing 32-bit data in SRAM compared to 16 or 8-bit data. However, data conversion is done either at the beginning or at the end of the forward pass in *BPLight-CNN*. Therefore, the effect is very minimal. Furthermore, it can also be noted from Fig. 12a and 12b that the speedup has a slightly decreasing trend from VGG-A to VGG-D. This is due to the increase in total number of convolution layers from VGG-A to VGG-D (VGG-A: 8 layers, VGG-B: 13 layers, VGG-C: 13 layers, VGG-D: 16 layers) as shown in Table 1. However, *BPLight-CNN*'s speedup for LeNet-A and LeNet-B are significantly higher than that for VGG-A though they have 9 and 10 layers, respectively. This increase in speedup for LeNet is due to its fewer number of WMAs. Lower number of WMAs means less computational time, which in turn leads to higher speedup.

Fig. 13a and 13b illustrate the comparison of computational efficiency (CE) (i.e., the total number of fixed-point operations performed per unit area in one second (GOPS/s/mm$^2$)) of the proposed *BPLight-CNN* with memristor crossbar based PipeLayer [4] and a baseline GPU based design. For training, GPU performs with an average CE of 305.67 GOPS/s/mm$^2$ (Min: 285, Max: 325); PipeLayer has an average CE of 2923 GOPS/s/mm$^2$(Min: 1710, Max: 3904). *BPLight-CNN*'s average CE during training is 99533.5 GOPS/s/mm$^2$ (Min: 58140, Max: 121024). Similarly, during inference GPU, PipeLayer, and *BPLight-CNN*'s average CEs are 130 GOPS/s/mm$^2$, 1425 GOPS/s/mm$^2$, and 44030 GOPS/s/mm$^2$ respectively. The proposed *BPLight-CNN* architecture shows a computational efficiency variance of 302 GOPS/s/mm$^2$ which is reasonable considering its high computational efficiency. PipeLayer uses memristor crossbars for the bulk of its arithmetic operations. Each memristor crossbar has a CE of 1707 GOPS. However, the overall CE of PipeLayer comes down to 1485 GOPS due to its extensive usage of data conversions. Also, ReLU and POOL are performed by a digital ALU in PipeLayer. This requires more memory to store intra-layer data for synchronizing with its pipeline mechanism. The superiority of *BPLight-CNN* comes from the fact that it is a completely analog accelerator. Therefore, *BPLight-CNN* does not involve inter-layer data conversions or storage for synchronization. AD and DA conversions are done either at the beginning or at the end of feature extraction in *BPLight-CNN*. On the contrary, PipeLayer's extensive inter-layer data conversions limits its computational efficiency. In addition to the compute efficient memristor, *BPLight-CNN* also uses high speed SOA as ReLU which has a CE in the order of 50000 GOPS/s/mm$^2$ [43]. In summary, compared to PipeLayer and GPU, *BPLight-CNN* has 34× and 325× higher computational efficiency during training, and 31× and 339× higher computational efficiency during inference. Weight resolution has a negligible effect on computational efficiency of *BPLight-CNN*, therefore, we do not present that result. Finally, *BPLight-CNN*'s computational efficiency has a decreasing trend similar to its speedup from VGG-A to VGG-D. This is due to the increase in computational requirements (with more convolution layers) from VGG-A to VGG-D. Also, *BPLight-CNN* has a significantly higher computational efficiency for LeNet-A and LeNet-B. This is due to the reduction in their memristive convolution (lower WMAs) requirements.

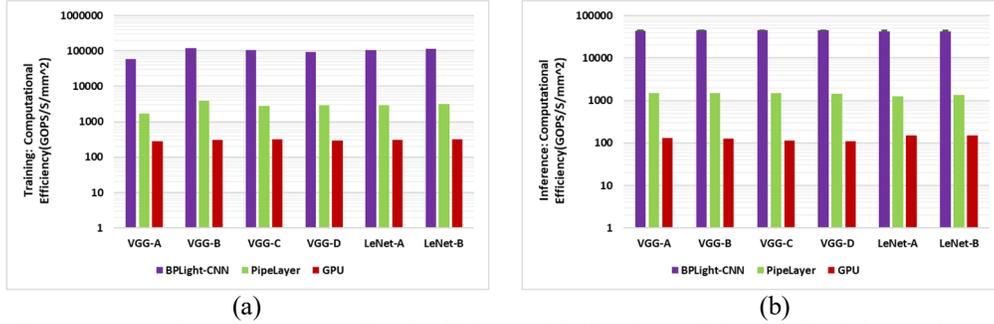

Fig.13: Comparison of computational efficiency [during (a) training & (b) inference] across accelerators (in GOPS/s/mm$^2$)

### 6.4 Energy Savings

We compare the energy efficiency of *BPLight-CNN* with PipeLayer and GPU as shown in Fig. 14a and 14b. During training, PipeLayer computes with an energy efficiency of 242.3 GOPS/s/W which is 11.6× higher than GPU based accelerator. *BPLight-CNN* outperforms both with an average energy efficiency of 9327.5 GOPS/s/W. PipeLayer replicates its early feature extraction layers several times (close to 50K times) to maintain a balanced pipeline. This involves excessive use of high-power consuming data conversions. *BPLight-CNN* uses passive optical components such as waveguides and comparators, in addition to energy efficient components such as ring modulators/demodulators, SOAs, and memristor. Also, *BPLight-CNN* uses very few ADCs/DACs compared to PipeLayer. As shown in Fig. 14(a), during training, we obtain 38.5× and 447× improvements in energy efficiency for *BPLight-CNN* compared to PipeLayer and GPU, respectively. Overall, the standard deviation of training energy efficiency of *BPLight-CNN* across benchmarks is 178 GOPS/s/mm$^2$. Similarly, Fig.14b demonstrates energy savings of *BPLight-CNN* as compared to PipeLayer and GPU during inference. *BPLight-CNN* improves energy-efficiency by 38.7× and 413× compared to PipeLayer and GPU.

Fig. 15a and 15b show the effects of weight resolution on overall energy efficiency during training and inference respectively, for all the benchmarks in *BPLight-CNN*. As the weight resolution increases, there is a minimal reduction in energy efficiency for both training and inference. During training, 32-bit energy efficiency is 4.5% lower compared to 16-bit efficiency, and 5.2% lower compared to 8-bit efficiency on average. During inference, 32-bit energy efficiency is 4.2% lower compared to 16-bit energy efficiency, and 7.4% lower compared to 8-bit energy efficiency, on average. The reduction in energy efficiency for high resolution is due to the increase in power consumption for high resolution DACs and ADCs. It can also be noted that there is a minimal increase in energy efficiency of *BPLight-CNN* from VGG-A to VGG-D. The increase in computational requirements from VGG-A to VGG-D improves utilization of all the components of *BPLight-CNN*, which results in an increase in energy efficiency. On the contrary, LeNet-A and LeNet-B have relatively less computational demand, which underutilizes our *BPLight-CNN* accelerator and leads to lower energy efficiency.

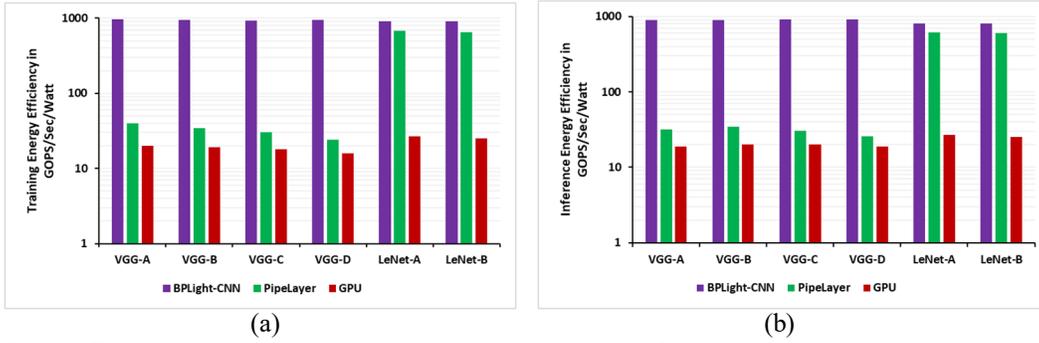

Fig.14: Comparison of energy efficiency [during (a) training & (b) inference] across accelerators (in GOPS/s/Watt)

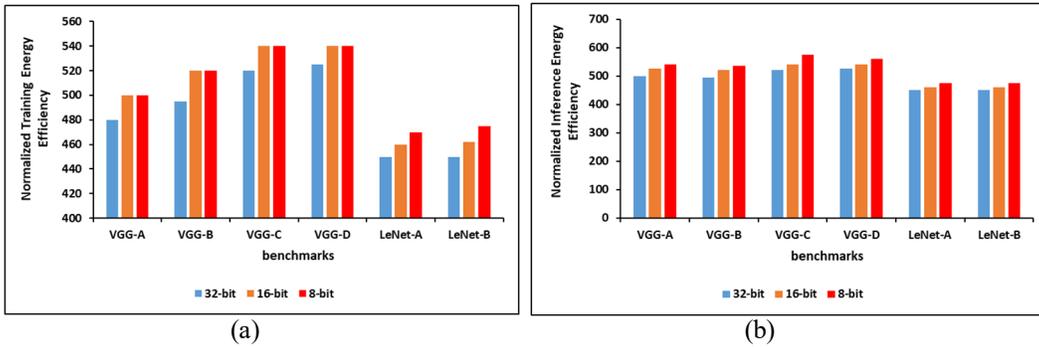

Fig.15: Normalized energy efficiency of *BPLight-CNN* w.r.t weight resolution (a) for training, and (b) for inference

Table 3. Results Summary

| Architecture | Speedup | Computational Efficiency | Energy Efficiency |
|---|---|---|---|
| *BPLight-CNN* | Highest | Highest | Highest |
| PipeLayer | Medium | Medium | Medium |
| GPU | Lowest | Lowest | Lowest |

## 6.5 Results Summary

We summarize the results presented in Section 6.3 and Section 6.4 in Table 3. From the table it is apparent that our novel *BPLight-CNN* accelerator outperforms previously proposed CNN accelerators by combining the photonics-based backpropagation accelerator with a configurable memristor-integrated photonic CNN accelerator design. The excellent performance and energy gains compared to previous approaches strongly motivate the use of *BPLight-CNN* to execute future CNN based workloads.

## 7. CONCLUSIONS

This work demonstrates a fully analog CNN accelerator called *BPLight-CNN* that integrates compute-efficient memristors and ultra-fast photonic components. We introduce a reconfigurable convolution design in each CNN layer to enable *BPLight-CNN* to emulate a range of sample CNN models. We also use a novel approach to handle analog signed-weight arithmetic in the memristive convolution layers. Compared to PipeLayer [4] and GPU implementations, the proposed *BPLight-CNN* architecture shows higher computational and energy efficiency due to the use of energy efficient SOAs, optical comparators, and also due to its use of a fully analog feature extraction method. We demonstrated that the proposed design has the potential to achieve (i) at least 34× acceleration in speedup, 34× improvement in computational efficiency, and 38.5× energy savings during training; and (ii) 29× acceleration in speedup, 31× improvement in computational efficiency, and 38.7× improvement in energy savings during inference compared to the state-of-the-art designs such as PipeLayer, at a 16-bit resolution. BPLight-CNN attains these performances with an approximately 6% reduction in accuracy over the state-of-the-art. The proposed architecture attains the state-of-the-art accuracy with a 32-bit resolution with negligible compromise in terms of power and throughput. Photonic components have insertion losses which may slightly affect the overall accuracy when the number of deep learning stages increases. Our future work will address the issue of broader applicability of our accelerator to other types of deep learning models, and explore the scalability of this architecture for larger neural network problem sizes.

## ACKNOWLEDGEMENTS

This work was supported by a postdoctoral fellowship from the American Association of Immunologists, and a doctoral teaching fellowship from the College of Engineering, Texas A&M University.

## REFERENCES

[1] C. Zhang, Z. Fang, P. Zhou, P. Pan, and J. Cong, "Caffeine: Towards Uniformed Representation and Acceleration for Deep Convolutional Neural Networks", *IEEE/ACM International Conference on Computer-Aided Design ICCAD, Nov. 2016*.

[2] A. Shafiee, A. Nag, N. Muralimanohar, R. Balasubramonian, J. P. Strachan, M. Hu, R. S. Williams, and V. Srikumar, "ISAAC: A Convolutional Neural Network Accelerator with In-Situ Analog Arithmetic in Crossbars", *IEEE/ACM International Symposium on Computer Architecture (ISCA), Jun. 2016*.

[3] T. Gokmen and Y. Vlasov "Acceleration of Deep Neural Network Training with Resistive Cross-Point Devices: Design Considerations", *Front. Neuroscience, July 2016*.

[4] L. Song, X. Qian, H. Li, and Y. Chen, "PipeLayer: A Pipelined ReRAM-Based Accelerator for Deep Learning", IEEE International Symposium on High Performance Computer Architecture (HPCA), Feb. 2017.

[5] T. Gokmen, O. M. Onen, and W. Haensch, "Training Deep Convolutional Neural Networks with Resistive Cross-Point Devices", *Frontier Neurosciences. Vol. 11, No. 538,* 2017.

[6] C. Li, D. Belkin, Y. Li, P. Yan, M. Hu, N. Ge, H. Jiang, E. Montgomery, P. Lin, Z. Wang, W. Song, J. P. Strachan, M. Barnell, Q. Wu, R. S. Williams, J. J. Yang, and Q. Xia, "Efficient and self-adaptive in-situ learning in multilayer memristor neural networks", *Nature Communications*, Vol. 9, No. 2385, 2018.

[7] Miao Hu, J. P. Strachan, Z. Li, E. M. Grafals, N. Davila, C. Graves, S. Lam, N. Ge, J. J. Yang, R. S. Williams, "Dot-Product Engine for Neuromorphic Computing: Programming 1T1M Crossbar to Accelerate Matrix-Vector Multiplication", *IEEE/ACM Design Automation Conference (DAC), 2016*.

[8] K Vandoorne, J. Dambre, D. Verstraeten, B. Schrauwen, and P. Bienstman, "Parallel Reservoir Computing Using Optical Amplifiers", *IEEE Transactions on Neural Networks, Vol. 22, No. 9, Sept. 2011, pp. 1469-1481*.

[9] A. N. Tait, T. Ferreira de Lima, E. Zhou, A. X. Wu, M. A. Nahmias, B. J. Shastr, and P. R. Prucnal, "Neuromorphic photonic networks using silicon photonic weight banks," in *Scientific Reports, Vol. 7, Art. No. 7430, 2017*.

[10] N. Janosik, Q. Cheng, M. Glick, Y. Huang, and K. Bergman, "High-resolution silicon microring based architecture for optical matrix multiplication," *in Conference on Lasers and Electro-Optics (CLEO), July 2019*.

[11] T. W. Hughes, M. Minkov, Y. Shi, and S. Fan, "Training of photonic neural networks through in situ backpropagation and gradient measurement," in Optica, Vol. 5, pp. 864-871, 2018.

[12] J. R. Ong, C. C. Ooi, T. Y. L. Ang, S. T. Lim, and C. E. Png, "Photonic Convolutional Neural Networks UsingIntegrated Diffractive Optics", IEEE Journal of Selected Topics In Quantum Electronics, Vol. 26, No. 5, September/October 2020.

[13] A. Mehrabian, Y. Al-Kabani, V. J. Sorger, and T. El-Ghazawi "PCNNA: A Photonic Convolutional Neural Network Accelerator", in Proceedings of IEEE SOCC 2018.

[14] Y. Shen, N. C. Harris, S. Skirlo, M. Prabhu, T. Baehr-Jones, M. Hochberg, X. Sun, S. Zhao, H. Larochelle, D. Englund, and M. Soljačić, "Deep learning with coherent nanophotonic circuits", *Nature Photonics, Vol. 11, pp. 441-446, Jun. 2017*.

[15] D Dang, J. Dass, and R. Mahapatra, "ConvLight: A Convolutional Accelerator with Memristor Integrated Photonic Computing", *IEEE International Conference on High Performance Computing (HiPC), Feb. 2017*.

[16] F. Zokaee, Q. Lou, N. Youngblood, W. Liu, Y. Xie and L. Jiang, "LightBulb: A Photonic-Nonvolatile-Memory-based Accelerator for Binarized Convolutional Neural Networks," *2020 Design, Automation & Test in Europe Conference & Exhibition (DATE)*, Grenoble, France, 2020, pp. 1438-1443, doi: 10.23919/DATE48585.2020.9116494.

[17] W. Liu, W. Liu, Y. Ye, Q. Lou, Y. Xie and L. Jiang, "HolyLight: A Nanophotonic Accelerator for Deep Learning in Data Centers," *2019 Design, Automation & Test in Europe Conference & Exhibition (DATE)*, Florence, Italy, 2019, pp. 1483-1488, doi: 10.23919/DATE.2019.8715195.

[18] Y. Long, L. Zhou, and Jian Wang, "Photonic-assisted microwave signal multiplication and modulation using a silicon Mach–Zehnder modulator", Scientific Reports, Vol. 6, Art. No. 20215, Feb. 2016.

[19] Semiconductor Optical Amplifiers, 2002 edition. *Boston; London: Springer, 2002.*

XX:24 • D. Dang et al.

[20] P. Li, X. Yi, X. Liu, D. Zhao, Y. Zhao, and Y. Wang, "All-optical Analog Comparator", *Nature Scientific Reports, Vol. 6, Art. No. 31903, Aug. 2016.*
[21] The MNIST Database. (2018). [online]. Available: http://yann.lecun.com/exdb/mnist/
[22] B. Dmitri, S. Strukov, R. Duncan, R Williams, "The Missing Memristor is found", *Nature 453, 80-83, 2008.*
[23] W. Bogaerts et. al, "Microring Resonators", *Laser Photonics Rev. 6, No. 1, 47–73, 2012.*
[24] J. M. Ramirez, Q. Liu, V. Vakarin, J. Frigerio, A. Ballabio, X. Le Roux, D. Bouville, L. Vivien, G. Isella, and D. Marris-Morini, "Graded SiGe waveguides with broadband low-loss propagation in the mid infrared", *Optics Express 2018.*
[25] S. Xiao, M. H. Khan, H. Shen, and M. Qi, "Modeling and measurement of losses in silicon-on-insulator resonators and bends," *Optic Express, Vol. 15, No. 17, pp. 553–✕1, 2007.*
[26] S. V. R. Chittamuru and S. Pasricha, "Crosstalk Mitigation for High-Radix and Low-Diameter Photonic NoC Architectures", *IEEE Design and Test, Vol. 32, No.3, 2015.*
[27] I. Thakkar, S. V. R. Chittamuru and S. Pasricha, "Run-Time Laser Power Management in Photonic NoCs with On-Chip Semiconductor Optical Amplifiers," *Proceedings of IEEE/ACM International Symposium on Networks-on-Chip (NOCS), Aug. 2016.*
[28] P. Grani and S. Bartolini, "Design Options for Optical Ring Interconnect in Future Client Devices," *ACM Journal on Emerging Technologies in Computing Systems (JETC), vol. 10, no. 4, 2014.*
[29] X. Xue, P. Wang, Y. Xuan, M. Qi, and A. M. Weiner, "Microresonator Kerr frequency combs with high conversion efficiency", Laser & Photonics Rev., vol.11, no.1, 2017.
[30] C. Li, M. Browning, P. V. Gratz, and S. Palermo, "Energy-efficient optical broadcast for nanophotonic networks-on-chip," *Optical Interconnects Conference, 64-65, 2012.*
[31] S. V. R. Chittamuru, D. Dang, S. Pasricha, and R. Mahapatra, "BiGNoC: Accelerating Big Data Computing with Application-Specific Photonic Network-on-Chip Architectures," *IEEE Transactions on Parallel and Distributed Systems, Vol. 29, no. 11, 2018.*
[32] J. J. Yang, D. B. Strukov, and D. R. Stewart, "Memristive Devices for Computing", *Nature Nanotechnology, Vol. 8, pp. 13-24, Dec. 2012.*
[33] C. Szegedy, W. Liu, Y. Jia, P. Sermanet, S. Reed, D. Anguelov, D. Erhan, V. Vanhoucke, and A. Rabinovich, "Going deeper with convolutions", *IEEE Conference on Computer Vision and Pattern Recognition (CVPR), 2015.*
[34] M. R. Uddin, T. K. Siang, N. Munarah, M. Norfauzi, N. Ahmed and M. A Salam, "Quality Analysis of a Photonic Micro-ring Resonator", *IEEE International Conference on Computer & Communication Engineering, 2016.*
[35] C. Sun, C. O. Chen, G. Kurian, L. Wei, J. Miller, A. Agarwal, L. Peh, and V. Stojanovic, "DSENT - A Tool Connecting Emerging Photonics with Electronics for Opto-Electronic Networks-on-Chip Modeling," in *IEEE/ACM International Symposium on Networks-on-Chip (NOCS), Lyngby, Denmark, May 2012.*
[36] Y. Ma, Z. Xuan, Y. Liu, R. Ding, Y. Li, A. Eu-Jin Lim, G. Lo, T. Baehr-Jones, and M. Hochberg, "Silicon microring based modulator and filter for high speed transmitters at 1310 nm," in *Optical Interconnects Conference*, San Diego, CA, 2014, pp. 23-24.
[37] T. Baba, S. Akiyama, M. Imai, N. Hirayama, H. Takahashi, Y. Noguchi, T. Horikawa, and T. Usuki, "50-Gb/s ring-resonator-based silicon modulator," in *Opt. Express, Vol. 21, pp. 11869-11876, 2013.*
[38] Y. Jia, E. Shelhamer, J. Donahue, S. Karayev, J. Long, R. Girshick, S. Guadarrama, and T. Darrell, "Caffe: Convolutional architecture for fast feature embedding", *ACM International Conference on Multimedia, Nov. 2014, pp. 675-678.*
[39] T. Fujita, Y. Toba, Y. Miyoshi, and M. Ohashi, "Optical Analog Multiplier based on Phase Sensitive Amplification", *OptoElectronics and Communications Conference (OECC), Sept. 2013.*
[40] A. Krizhevsky, I. Sutskever, and G. E. Hinton, "ImageNet Classification with Deep Convolutional Neural Network", *International Conference on Neural Information Processing Systems (NIPS), Dec. 2012.*
[41] M. Saberi, R. Lotfi, K. Mafinezhad and W. A. Serdijn, "Analysis of Power Consumption and Linearity in Capacitive Digital-to-Analog Converters Used in Successive Approximation ADCs," in *IEEE Transactions on Circuits and Systems I: Regular Papers*, vol. 58, no. 8, pp. 1736-1748, Aug. 2011.
[42] K. Simonyan and A. Zisserman, "Very Deep Convolutional Networks for Large-scale Image Recognition", *International Conference on Learning Representations (ICLR), 2015.*
[43] Y. Lecun, L. Bottou, Y. Bengio, and P. Haffner, "Gradient-based learning applied to document recognition", *Proceedings of the IEEE, Vol. 86, No. 11 Nov 1998, pp. 2278-2324.*
[44] A. Khorami, M. B. Dastjerdi and A. F. Ahmadi, "A low-power high-speed comparator for analog to digital converters," in IEEE International Symposium on Circuits and Systems (ISCAS), Montreal, pp. 2010-2013, 2016.
[45] O. Russakovsky, J. Deng, H. Su, J. Krause, S. Satheesh, S. Ma, Z. Huang, A. Karpathy, A. Khosla, M. Bernstein, A. C. Berg, and L. Fei-Fei, "ImageNet Large Scale Visual Recognition Challenge", *International Journal of Computer Vision (IJCV), Vol. 115, No. 3, Dec. 2015, pp. 211-252.*
[46] IPKISS-Photonic Framework. (2018) [online]. Available: www.lucedaphotonics.com
[47] D. Dang, "Energy-Efficient Photonic Architectures for Large-Scale Data Analytics", Doctoral dissertation, Texas A & M University, 2018. Available electronically from http : / /hdl .handle .net /1969 .1 /174537.
[48] S. Pasricha, S. V. R. Chittamuru, I. G Thakkar, and V. Bhat, "Securing photonic NoC architectures from hardware trojans", in Proceedings IEEE/ACM International Symposium on Networks-on-Chip (NOCS), pp 1-8, 2018.
[49] S. Pasricha, S. V. R. Chittamuru, I. G Thakkar "Cross-Layer Thermal Reliability Management in Silicon Photonic Networks-on-Chip", Proceedings of Great Lakes Symposium on VLSI, pp 317-322, 2018.
[50] S. V. R. Chittamuru, I. G. Thakkar, S. Pasricha, S. S. Vatsavai and V. Bhat, "Exploiting Process Variations to Secure

Photonic NoC Architectures from Snooping Attacks," in *IEEE Transactions on Computer-Aided Design of Integrated Circuits and Systems*, doi: 10.1109/TCAD.2020.3014184.
[51]  D. Dang, S. V. R. Chittamuru, R. Mahapatra and S. Pasricha, "Islands of heaters: A novel thermal management framework for photonic NoCs," in proceedings of IEEE/ACM ASP-DAC, Chiba, 2017, pp. 306-311.